\documentclass[a4paper, 10pt, conference]{ieeeconf}      
\usepackage{FG2024}
\usepackage{cite}
\usepackage{amsmath,amssymb,amsfonts}
\usepackage{algorithmic}
\usepackage{graphicx}
\FGfinalcopy 
\usepackage{url}
\IEEEoverridecommandlockouts                              
\title{\LARGE \bf
Dynamic Cross Attention for Audio-Visual Person Verification
}

\author{\parbox{16cm}{\centering
    {\large R. Gnana Praveen and Jahangir Alam}\\
    {\normalsize
    Computer Research Institute of Montreal (CRIM), Canada}}
    \thanks{The authors wish to acknowledge the funding from the Government of Canada’s New Frontiers in Research Fund (NFRF) through grant NFRFR-2021-00338.}
}

\begin{document}

\ifFGfinal
\thispagestyle{empty}
\pagestyle{empty}
\else
\author{Anonymous FG2024 submission\\ Paper ID 308 \\}
\pagestyle{plain}
\fi
\maketitle

\begin{abstract}

Although person or identity verification has been predominantly explored using individual modalities such as face and voice, audio-visual fusion has recently shown immense potential to outperform unimodal approaches. Audio and visual modalities are often expected to pose strong complementary relationships, which plays a crucial role for effective audio-visual fusion. However, they may not always strongly complement each other, they may also exhibit weak complementary relationships, resulting in poor audio-visual feature representations. In this paper, we propose a Dynamic Cross Attention (DCA) model that can dynamically select the cross-attended or unattended features on the fly based on the strong or weak complementary relationships, respectively, across audio and visual modalities. In particular, a conditional gating layer is designed to evaluate the contribution of the cross-attention mechanism and choose cross-attended features only when they exhibit strong complementary relationships, otherwise unattended features. 
Extensive experiments are conducted on the Voxceleb1 dataset to demonstrate the robustness of the proposed model. 
Results indicate that the proposed model consistently improves the performance on multiple variants of cross-attention while outperforming the state-of-the-art methods. Code is available at 
\url{https://github.com/praveena2j/DCAforPersonVerification}    
\end{abstract}

\section{INTRODUCTION}

Person Verification (PV) is a hot research topic in the field of biometrics, spanning a wide range of applications such as forensics, commercial and law enforcement applications \cite{7298570}. The voice and face are two predominant contact-free channels, widely explored for the task of verifying the identity of a person \cite{9442674,WANG2021215}. With the advancement of sophisticated deep learning architectures \cite{WANG2021215,6909616} and loss functions \cite{8953658,10.1007/978-3-319-46478-7_31,8578650}, both face and speaker verification systems have achieved remarkable success in improving the performance of PV. Despite the success of individual modalities, the performance of the system is still limited by the quality of the facial images or speech signals. For instance, when the facial images are corrupted by extreme pose variations, blur or low illumination, and speech signals by background noise or interference of other signals, the performance of the system will deteriorate. Therefore, audio-visual (A-V) fusion has been recently gaining a lot of attention as they are often expected to complement each other, which plays a crucial role in outperforming unimodal approaches \cite{10095234, praveen2023audiovisual}.

\begin{figure}[!t]
\centering
\includegraphics[width=1.0\linewidth]{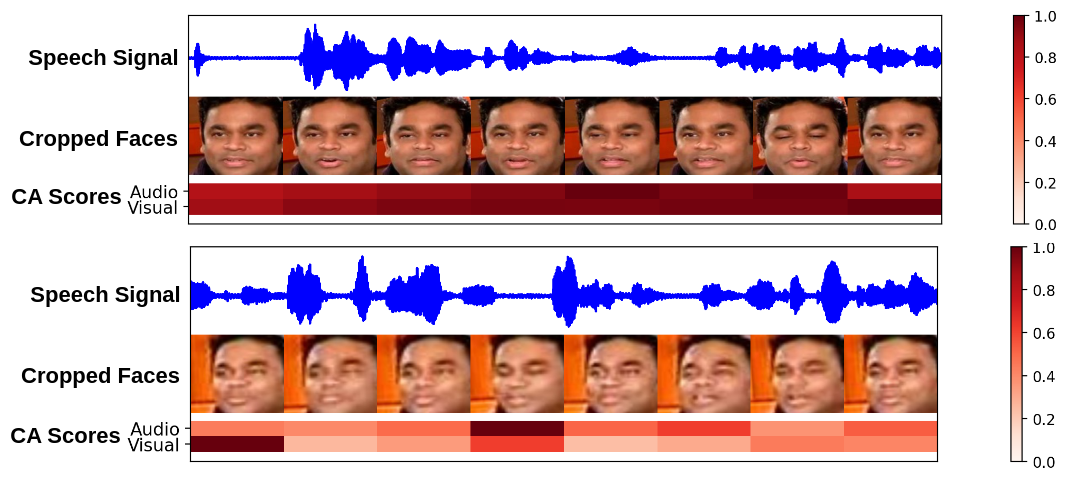}
\caption{\small Attention scores based on cross-attention. In the top image, both audio and visual modalities strongly complement each other, thereby assigning higher attention scores for face and voice. In the bottom image, the facial modality is corrupted due to blur, however, speech signal is not corrupted. Attending the corrupted face to uncorrupted speech signal fails to assign higher attention scores for speech signals.}
\label{Demo of Complementary relationships}
\vspace{-3mm}
\end{figure}

Recently, the performance of multimodal fusion has been significantly boosted by leveraging the complementary relationships across the modalities \cite{9758834,9667055}. Cross Attention (CA) is one of the widely used approaches to effectively capture complementary relationships across the modalities, which has been successfully explored in several applications such as action localization \cite{lee2021crossattentional}, emotion recognition \cite{9667055,10005783} and person verification \cite{praveen2023audiovisual}. The idea of CA is to leverage the complementary relationships by exploiting the knowledge of one modality to attend to another modality \cite{10123038}. However, audio and visual modalities may not always exhibit strong complementary relationships, they may also exhibit weak complementary relationships \cite{9706879,praveen2024cross}. Wang et al. \cite{9552921} provided a visual interpretability analysis of A-V fusion and demonstrated that audio and visual modalities may also pose conflicting (when one of the modalities is noisy) and dominating relationships (when one of the modalities is restrained). When one of the modalities is noisy or restrained due to background noise or clutter, leveraging the noisy or weak modality to attend to the good modality will degrade the features of uncorrupted (good) modality also, resulting in poor A-V feature representation as shown in Fig \ref{Demo of Complementary relationships}. 
Therefore, it is important to decide when or how to integrate the information from multiple modalities for effective A-V fusion based on their relevance for accurate PV. 
In this work, we have investigated the prospect of developing an adaptive A-V fusion robust to weak complementary relationships, while still retaining the potential of strong complementary relationships. To address the problem of weak complementary relationships, we propose a Dynamic Cross Attention (DCA) model to dynamically adapt to the strong and weak complementary relationships across the audio and visual modalities by choosing the most relevant features. Specifically, we introduce a conditional gating layer for each modality to evaluate the contribution of the CA mechanism based on the dependency on the other modality and select the cross-attended features only when they exhibit strong complementary relationships, otherwise unattended features. Therefore, the proposed DCA model adds more flexibility to the CA framework and improves the fusion performance even when the modalities exhibit weak complementary relationships. 
The proposed model 
can also be adapted to other variants of the CA model, thereby proving the robustness of the proposed model. 
The major contributions of the paper can be summarized as follows. (1) To our knowledge, this is the first work to investigate weak complementary issues across audio and visual modalities for PV. (2) We propose a DCA model to adaptively choose the cross-attended or unattended features to effectively capture the inter-modal relationships across audio and visual modalities. (3) Extensive experiments were conducted on the Voxceleb1 dataset and showed that the proposed model achieves consistent improvement over multiple variants of CA while outperforming state-of-the-art methods.



\section{Related Work}

Most of the existing approaches based on A-V fusion for PV either rely on early feature-level fusion \cite{Chen2020,9350195} or score-level fusion \cite{251181,FITPUB12292}.  
Sari et al. \cite{9414260} explored a common representation space to perform cross-modal verification using a shared classifier for both modalities. Shon et al. \cite{8683477} proposed a noise-tolerant attention mechanism to conditionally select the salient modality among audio and visual representations to deal with the problem of noisy modalities. Hormann et al. \cite{9320237} further improved the idea of \cite{8683477} 
by combining audio and visual features at intermediate layers, thus improving the performance of the system.
Chen et al. \cite{Chen2020} investigated various fusion strategies at the embedding level and showed that gating-based fusion outperforms other fusion strategies such as conventional soft attention and compact bilinear pooling. 

Though most of the prior approaches explored complementary relationships to deal with the problem of noisy samples, they failed to effectively capture the complementary relationships using cross-modal interactions.
Recently, Mocanu et al. \cite{9922810} explored CA based on cross-correlation across the audio and visual modalities to effectively capture the complementary relationships. Liu et al. \cite{10095883} explored cross-modal attention by deploying cross-modal boosters in a pseudo-siamese structure to model one modality by exploiting the knowledge from another modality. However, they focus only on inter-modal relationships \cite{9922810} or capture the intra- and inter-modal relationships in a decoupled fashion \cite{10095883}. Praveen et al. \cite{praveen2023audiovisual} explored a joint cross-attentional framework to simultaneously capture the intra- and inter-modal relationships by introducing joint feature representation in the CA framework. 
Most of these approaches assume that audio and visual modalities exhibit strong complementary relationships, thereby declining performance when they pose weak complementary relationships. 



\section{Proposed Approach}
\label{sec:pagestyle}
\noindent \textbf{A) Notations:}
For an input video sub-sequence $S$, $L$ non-overlapping video clips are uniformly sampled, and the corresponding deep feature vectors are obtained from the pre-trained models of audio and visual modalities. Let ${\boldsymbol X}_{a}$ and ${\boldsymbol X}_{v}$ denote the deep feature vectors of audio and visual modalities respectively for the given input video sub-sequence $S$ of fixed size, which is expressed as 
${ \boldsymbol X}_{a}=  \{  \boldsymbol {x}_{a}^1,  \boldsymbol x_{a}^2, ...,  \boldsymbol x_{ a}^L \} \in \mathbb{R}^{d_a\times L}$ 
and 
${ \boldsymbol X}_{\mathbf v}=  \{  \boldsymbol x_{ v}^1,  \boldsymbol x_{ v}^2, ...,  \boldsymbol x_{ v}^L \} \in \mathbb{R}^{d_v\times L}$
where ${d_a}$ and ${d_v}$ represent the dimensions of the audio and visual feature vectors, respectively, $ \boldsymbol x_{ a}^{ l}$ and $ \boldsymbol x_{ v}^{ l}$ denotes the audio and visual feature vectors of the video clips, respectively, for $l = 1, 2, ..., L$ clips.

\begin{figure*}[!t]
\centering
\includegraphics[width=0.85\linewidth]{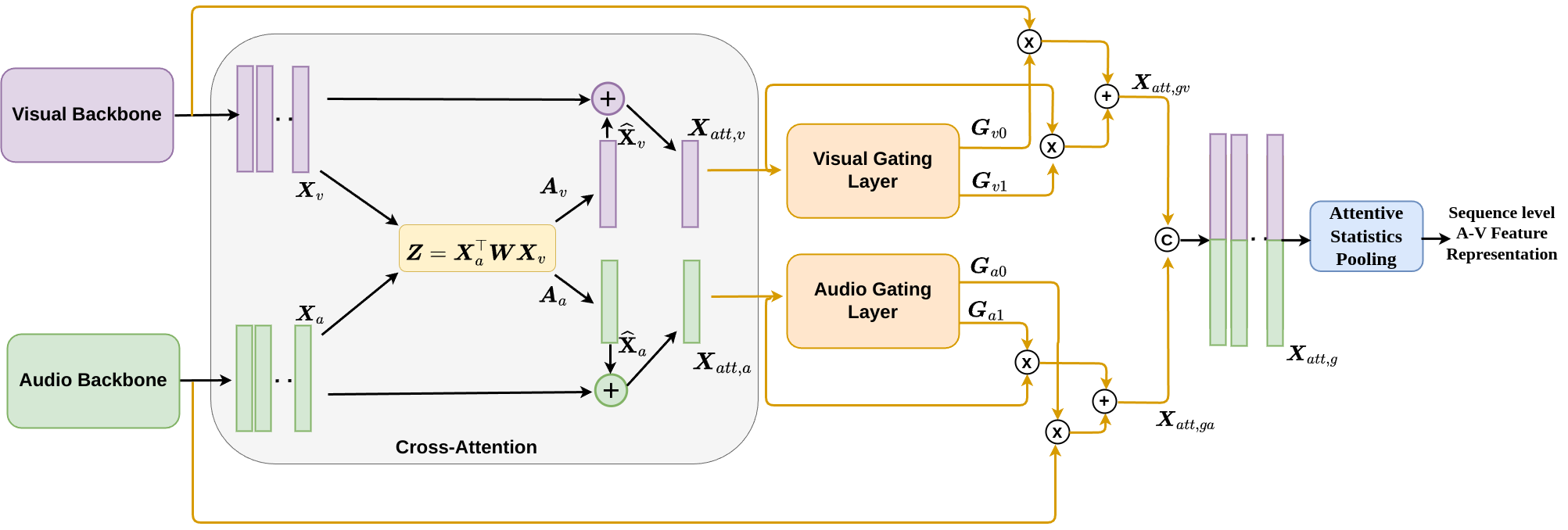}
\caption{Illustration of the proposed Dynamic Cross-Attention (DCA) model with vanilla Cross-Attention (CA) as the baseline.}
\label{AdapJA}
\vspace{-3mm}
\end{figure*}

\noindent \textbf{B) Preliminary - Cross Attention:}
In this work, we used vanilla CA \cite{9922810} as a baseline model for the proposed approach as shown in Fig \ref{AdapJA}. As a preliminary, we briefly present our baseline fusion model, CA \cite{9922810}.   
To capture the inter-modal relationships, cross-correlation is computed across the audio and visual modalities as
$\boldsymbol Z={\boldsymbol X}_{ a}^\top \boldsymbol W{\boldsymbol X}_{ v}$,
where $\boldsymbol Z\in\mathbb{R}^{L\times L}$, $\boldsymbol W\in\mathbb{R}^{d_a\times d_v}$ represents cross-correlation weights among audio and visual features. 
The high correlation coefficient of the cross-correlation matrix $\boldsymbol Z$ shows that the corresponding audio and visual features are highly related to each other. Based on this correlation matrix, CA weights of audio and visual features are computed 
by applying column-wise softmax of $\boldsymbol Z$ and $\boldsymbol Z^\top$, respectively as
\begin{align}
{\boldsymbol A}_{{ a}_{i,j}}
=\frac{ e^{{\mathbf Z}_{i,j}}}{\overset{ K}{\underset{ k\boldsymbol=1}{\sum}} e^{{\mathbf Z}_{k,j}}}    
\quad  \text{and} \quad
{\boldsymbol A}_{{ v}_{i,j}}
=\frac{ e^{{\mathbf Z^\top}_{i,j}}}{\overset{K}{\underset{ k\boldsymbol=1}{\sum}} e^{{\mathbf Z^\top}_{i,k}}}    
\end{align}
Now the CA weights of the individual modalities are used to modulate the corresponding feature representations to obtain more discriminative attention maps, which are given by
\begin{align}
\widehat{\mathbf{X}_{a}}=\mathbf{X}_{a}{{\mathbf A}_{ a}} \quad  \text{and}  \quad
\widehat{\mathbf{X}_v}=\mathbf X_v{{\mathbf A}_{ v}}
\end{align}
Finally, the cross-attended features of the individual modalities are obtained 
by adding the attention maps to the corresponding feature representations as
\begin{align}
{\boldsymbol X}_{att,a}= \tanh({\boldsymbol X}_{ a}+\widehat{\mathbf X_a}) \\ 
{\boldsymbol X}_{att,v}= \tanh({\boldsymbol X}_ v+\widehat{\mathbf X_v})
\end{align}
\noindent \textbf{C) Dynamic Cross Attention (DCA):}
In order to control the impact of each modality on the other modality and adaptively fuse the audio and visual modalities,  
we design a conditional gating layer using a fully connected layer for each modality separately. The objective of the conditional gating layer is to evaluate the importance of the cross-attended features by estimating the attention scores for the cross-attended and unattended features, as given by
\begin{align}
{\boldsymbol Y}_{go,v} = {\boldsymbol X}_{ att,v}^\top \boldsymbol W_{gl,v}
\hspace{-2mm} \quad \text{and} \hspace{-2mm} \quad 
{\boldsymbol Y}_{go,a} = {\boldsymbol X}_{att,a}^\top \boldsymbol W_{gl,a}
\end{align}
where $\boldsymbol W_{gl,a} \in\mathbb{R}^{d_a\times 2}$, $\boldsymbol W_{gl,v}\in\mathbb{R}^{d_v\times 2}$ are the learnable weights of the gating layers and $\boldsymbol Y_{go,a} \in\mathbb{R}^{L\times 2}$, $\boldsymbol Y_{go,v} \in\mathbb{R}^{L\times 2}$ are outputs of gating layers of audio and visual modalities respectively. The output of the gating layer is activated using a soft-max function with a small temperature $T$\cite{NEURIPS2019_f1748d6b}, to derive probabilistic attention scores, as given by 
\begin{align}
{\boldsymbol G}_{{a}}
=\frac{ e^{{\boldsymbol Y}_{go,a}/T}}{\overset{ K}{\underset{ k\boldsymbol=1}{\sum}} e^{{\boldsymbol Y}_{go,a}/T}}    
\quad  \text{and} \quad
{\boldsymbol G}_{{v}}
=\frac{ e^{{\boldsymbol Y}_{go,v}/T}}{\overset{K}{\underset{ k\boldsymbol=1}{\sum}} e^{{\boldsymbol Y}_{go,v}/T}}    
\end{align}
where $\boldsymbol {G}_{a} \in\mathbb{R}^{L\times 2}, \boldsymbol {G}_{v}\in\mathbb{R}^{L\times 2}$ denotes the probabilistic attention scores of audio and visual modalities respectively. $K$ denotes the number of output units of the gating layer, which is $2$, one for cross-attended features and one for unattended features.  

Based on these probabilistic attention scores, we can be able to estimate the relevance of cross-attended or unattended features depending on strong or weak complementary relationships respectively across the modalities. 
Ideally, for strong complementary relationships, the gating output of the cross-attended features (selected) will be $1$ and $0$ for unattended features (non-selected), which is the same as vanilla cross-attention \cite{9922810} and vice-versa. Additionally, we allowed a small weightage for the non-selected features in order to provide a regularization effect \cite{10.5555/3295222.3295264}. Empirically, we have set the value of $T$ to 0.1 in our experiments. By choosing a small value for $T$, the non-selected feature, which acts as a noisy signal helps to improve the generalization capability of the proposed model, thereby providing a regularization effect \cite{10.5555/3295222.3295264}. 
The two columns of $\boldsymbol G_a$ refer to the probabilistic attention scores of unattended features (first column) and cross-attended features (second column). To further multiply these attention scores with the corresponding feature representations, each column of the gating outputs is replicated to match the dimension of the corresponding feature vectors, denoted by 
$\boldsymbol G_{a0}$, $\boldsymbol G_{a1}$ and $\boldsymbol G_{v0}$, $\boldsymbol G_{v1}$ for audio and visual modalities respectively. 
The replicated attention scores are further multiplied with the corresponding cross-attended and unattended features of the respective modalities, followed by ReLU activation function to obtain the final attended features, which is given by
\begin{align}
{\boldsymbol X}_{att,gv} = \text{ReLU}(\boldsymbol X_{v} \otimes \boldsymbol G_{v0} + {\boldsymbol X}_{att,v} \otimes \boldsymbol G_{v1})  \\
{\boldsymbol X}_{att,ga} = \text{ReLU}(\boldsymbol X_{a} \otimes \boldsymbol G_{a0} + {\boldsymbol X}_{att,a} \otimes \boldsymbol G_{a1})      
\end{align}

where $\otimes$ denotes element-wise multiplication. ${\boldsymbol X}_{att,ga}, {\boldsymbol X}_{att,gv}$ denote the final attended feature vectors of audio and visual modalities respectively obtained from the DCA model. The final attended audio and visual representations obtained from the proposed DCA model are fed to the attentive statistics pooling (ASP) \cite{okabe18_interspeech} to obtain utterance-level representations of A-V feature vectors. Finally, the cosine distance similarity scores are obtained from the utterance-level A-V representations, where the parameters of the proposed model along with the ASP module are optimized using Additive Angular Margin Softmax (AAMSoftmax) loss function\cite{8953658}.


\section{Results and Discussion}
\noindent \textbf{A) Dataset:} 
The proposed approach has been evaluated on the Voxceleb1 dataset \cite{Nagrani17}, captured under challenging environments from YouTube videos. The dataset has 1,48,642 video clips, from 1251 speakers of different ethnicities, accents, professions, and ages. The dataset is gender balanced with 55\% of speakers being male, and the duration of each video clip is 4 to 145 seconds. Out of 1251 speakers, 1211 speakers are partitioned as development set and 40 speakers as test set (Vox1-O). In our experimental framework, we split the Voxceleb1 development set into 1150 and 61 speakers as training and validation sets respectively. The results are reported on both validation split and Vox1-O test sets. 

\noindent \textbf{B) Evaluation Metrics:}
The performance of the proposed approach is measured using Equal Error Rate (EER) and minimum Detection Cost Function (minDCF), which has been widely used for speaker verification in the literature \cite{9922810,9889705}. EER refers to the point in Detection Error Tradeoff (DET) curve, where the False Accept Rate (FAR) is equal to the False Reject Rate (FRR). So lower EER indicates better performance and high reliability of the system. DCF provides the control for the costs associated with false alarms (false positives) and missed detections (false negatives) \cite{BRUMMER2006230}. In our experiments, we considered the parameters of the DCF as $P_{target}=0.05$, $C_{miss}=1$ and $C_{falsealarm}=1$ similar to that of VoxSRC-20 \cite{nagrani2020voxsrc}.        

\noindent \textbf{C) Ablation Study:}
We reported the results based on the average of three runs for statistical stability. The audio and visual feature representations are extracted using ECAPA-TDNN \cite{Desplanques2020} and Resnet-18 \cite{7780459} respectively similar to that of \cite{10096814}. 
We compared the performance of the proposed DCA model with some of the widely used fusion strategies as shown in Table \ref{tab1}. First, we performed an experiment of score-level fusion, where the similarity scores are obtained from individual modalities and then fused. Next, we implemented early feature-level fusion, where audio and visual features are concatenated and subsequently used to obtain similarity scores. We have observed that the performance of early feature fusion was better than that of score-level fusion due to the fusion of low-level information across the modalities. We further explore some of the relevant attention mechanisms that are widely used in the literature. We used fusion with self-attention, where the concatenated audio and visual features are fed to the self-attention module. The fusion performance of self-attention has been improved over the prior two strategies as they leverage the temporal dynamics to obtain semantic A-V feature representations. 

Next, we implemented CA to capture the inter-modal relationships based on cross-correlation across the audio and visual modalities. Since inter-modal relationships play a crucial role in leveraging the efficacy of A-V fusion, the performance of CA is found to be better than that of prior fusion strategies. Additionally, we also explored joint cross-attention (JCA) \cite{praveen2023audiovisual}, which further improves the fusion performance by introducing the joint A-V feature representation in the CA framework to simultaneously capture the intra- and inter-modal relationships. Since the proposed DCA model adds flexibility to the CA framework to deal with weak complementary relationships, we have evaluated the performance of the proposed model on both variants of CA: CA \cite{9922810} and JCA \cite{praveen2023audiovisual}. We can observe that by employing the proposed DCA model, the fusion performance has been consistently improved for both CA \cite{9922810} and JCA \cite{praveen2023audiovisual} by handling the weak complementary relationships. The performance boost of the proposed model is more emphasized in CA \cite{9922810} than JCA \cite{praveen2023audiovisual} with a relative improvement of 9.3\% for CA and 2.9\% for JCA in terms of EER (similar trend of improvement for minDCF also). We hypothesize that since JCA depends on both intra- and inter-modal relationships, the problem of weak complementary inter-modal relationships is less pronounced in JCA than in CA.  
Finally, we also included the impact of DCA on JCA with BLSTMs as in \cite{praveen2023audiovisual}.   

\begin{table}
\centering
\caption{Performance of various fusion strategies on the validation set}\label{tab1}
\begin{tabular}{|c|c|c|}
\hline
\textbf{Fusion} &   \multicolumn{2}{|c|}{\textbf{Validation Set}} \\  
\cline{2-3}
\textbf{Method} &   \textbf{EER $\downarrow$} &  \textbf{minDCF $\downarrow$} \\
\hline
\hline
Score-level Fusion &  2.521 & 0.217 \\ \hline
Feature Concatenation  & 2.489 & 0.193 \\ \hline
Self-Attention &  2.412 & 0.176\\ \hline
Cross-Attention (CA) &  2.387 & 0.149 \\ \hline
Joint Cross-Attention (JCA) &  2.315 & 0.135\\ \hline \hline
CA + DCA & 2.166  & 0.132 \\ \hline
JCA + DCA & 2.247& 0.127\\
\hline
 JCA (w/ BLSTMs) + DCA& \textbf{2.138}&\textbf{0.119}\\ \hline
\end{tabular}
\vspace{-3mm}
\end{table}

\begin{table}
\scriptsize
\centering
\caption{Performance of the proposed approach in comparison to state-of-the-art on the validation set and Vox1-O set }\label{tab3}
\begin{tabular}{|c|c|c|c|c|c|c|c|c|c|c|}
\hline
\textbf{Fusion} &   \multicolumn{2}{|c|}{\textbf{Validation Set}} & \multicolumn{2}{|c|}{\textbf{Vox1-O Set}} \\
\cline{2-5}
\textbf{Method} &  \textbf{EER $\downarrow$} &  \textbf{minDCF $\downarrow$} & \textbf{EER $\downarrow$} &  \textbf{minDCF $\downarrow$} \\
\hline
\hline
Visual & 3.720 & 0.298 & 3.779 & 0.274 \\ \hline
Audio & 2.553 & 0.253 & 2.529  & 0.228 \\ \hline
Tao et al \cite{10096814} &  2.476 &  0.203 & 2.409 & 0.198\\ \hline
Chen et al \cite{Chen2020} & 2.403  & 0.163 & 2.376  & 0.161 \\ \hline

Mocanu et al \cite{9922810} & 2.387  & 0.149 & 2.355 & 0.156 \\ \hline
 Praveen et al \cite{praveen2023audiovisual} & 2.173  & 0.126 & 2.214 & 0.129 \\ \hline
Mocanu et al \cite{9922810}+DCA & 2.166 & 0.132 & 2.193 & 0.145 \\
\hline
Praveen et al \cite{praveen2023audiovisual}+DCA & \textbf{2.138}  & \textbf{0.119} & \textbf{2.172} & \textbf{0.121} \\
\hline
\end{tabular}
\vspace{-3mm}
\end{table}

\noindent \textbf{D) Comparison to state-of-the-art:}
The proposed approach has been compared with other state-of-the-art methods by training the models on the Voxceleb1 development set as shown in Table \ref{tab3}. Since a majority of the existing methods have been trained on the Voxceleb2 development set using different experimental protocols, we follow the experimental setup of \cite{10096814} to have a fair comparison. In particular, we have used the cleansed samples obtained using the approach of \cite{10096814} to train the models of our approach as well as other relevant state-of-the-art methods \cite{9922810,Chen2020,praveen2023audiovisual}.
We also evaluated the performance of the individual modalities and found that the audio modality performs relatively better than visual modality. Tao et al. \cite{10096814} explored complementary relationships across audio and visual modalities to discriminate clean and noisy samples. Chen et al. \cite{Chen2020} further improved performance by controlling the flow of information across the modalities depending on their importance. Mocanu et al. \cite{9922810} and Praveen et al. \cite{praveen2023audiovisual} explored CA frameworks and showed better performance than \cite{Chen2020} and \cite{10096814}. Praveen et al. \cite{praveen2023audiovisual} demonstrated better performance than that of \cite{9922810} by deploying joint feature representation in the CA framework to simultaneously capture both intra- and inter-modal relationships. 
Since the proposed model improves the performance of cross-modal attention by handling weak complementary relationships, we have compared DCA with both \cite{9922810} and \cite{praveen2023audiovisual}. Though \cite{Chen2020} also explored gating mechanisms, they focused on controlling the flow of information across the modalities. However, we have used a gating mechanism to control the flow of cross-attended and unattended features to deal with weak complementary relationships within the modalities. 
We can observe that the proposed DCA model consistently boosts the performance of both variants of CA: CA \cite{9922810} and JCA \cite{praveen2023audiovisual}. 

\section{Conclusion}

In this paper, we investigated the issues with weak complementary relationships across the audio and visual modalities for PV. Although CA-based approaches have shown significant improvement in the fusion performance, weak complementary relationships often degrade the fusion performance by deteriorating the fused A-V feature representations. 
To address this issue, we proposed a simple, yet efficient DCA model to effectively capture the inter-modal relationships by handling the problem of weak complementary relationships, while retaining the benefit of strong complementary relationships.
The performance of the proposed approach can be further enhanced by training with the large-scale Voxceleb2 dataset as it can improve the generalization ability. 




{\small
\bibliographystyle{ieeetr}
\bibliography{egbib.bib}
}

\end{document}